\pdfoutput=1
\documentclass[acmsmall]{acmart}

\AtBeginDocument{%
  \providecommand\BibTeX{{%
    \normalfont B\kern-0.5em{\scshape i\kern-0.25em b}\kern-0.8em\TeX}}}

\setcopyright{acmcopyright}
\copyrightyear{2023}
\acmYear{2023}

\usepackage{hyperref}
\usepackage{listings}
\usepackage{color}
\usepackage{xcolor}
\usepackage[normalem]{ulem}
\usepackage{makecell}
\definecolor{codegreen}{rgb}{0,0.6,0}
\definecolor{codegray}{rgb}{0.5,0.5,0.5}
\definecolor{codepurple}{rgb}{0.58,0,0.82}
\definecolor{backcolour}{rgb}{0.95,0.95,0.92}
\usepackage{multirow}

\usepackage{amssymb}

\newcommand{\add}[1]{{#1}}
\newcommand{\delete}[1]{} 
\newcommand{\replace}[2]{\delete{#1}\add{#2}}

\begin{document}

\title{Online Training of Large Language Models: Learn while Chatting}

\author{Juhao Liang}
\authornote{Both authors contributed equally to this research.}
\email{juhaoliang1997@gmail.com}
\author{Ziwei Wang}
\authornotemark[1]
\email{ziweiwang2@link.cuhk.edu.cn}
\author{Zhuoheng Ma}
\authornotemark[1]
\email{zhuohengma@link.cuhk.edu.cn}  
\affiliation{%
  \institution{The Chinese University of Hong Kong, Shenzhen}
  \country{China}
}

\author{Jianquan Li}
\author{Zhiyi Zhang}
\author{Xiangbo Wu}
\affiliation{%
  \institution{The Chinese University of Hong Kong, Shenzhen}
  \country{China}
}




\author{Benyou Wang}
\affiliation{%
  \institution{The Chinese University of Hong Kong, Shenzhen \& Shenzhen Research Institute of Big Data}
  \country{China}
}
\email{wangbenyou@cuhk.edu.cn}


\begin{abstract}

Large Language Models (LLMs) have dramatically revolutionized the field of Natural Language Processing (NLP), offering remarkable capabilities that have garnered widespread usage. However, existing interaction paradigms between LLMs and users are constrained by either inflexibility, limitations in customization, or a lack of persistent learning. This inflexibility is particularly evident as users, especially those without programming skills, have restricted avenues to enhance or personalize the model. Existing frameworks further complicate the model training and deployment process due to their computational inefficiencies and lack of user-friendly interfaces. To overcome these challenges, this paper introduces a novel interaction paradigm-'Online Training using External Interactions'-that merges the benefits of persistent, real-time model updates with the flexibility for individual customization through external interactions such as AI agents or online/offline knowledge bases \footnote{\url{https://github.com/FreedomIntelligence/Online-Training.git}}.

\end{abstract}




\delete{\received{20 February 2007}}

\maketitle

\section{Introduction}

Large language models~(LLMs)~\cite{shanahan2022talking, wei2022chain, hoffmann2022training, taylor2022galactica} has witnessed remarkable advancements in recent years, revolutionizing various natural language processing~(NLP) tasks~\cite{brants2007large, bahl1989tree, collobert2011natural}. 
In a world where knowledge and user requirements are constantly shifting, it's critical for these models to engage in incremental  learning~\cite{zhao2023survey} . Existing work  ~\cite{huang2022large,shinn2023reflexion} tried to improve language models by self-consistency or \replace{Self-Reflection}{self-reflection}; however such improvement is limited to be in the scenarios where deterministic rule based check could be used like coding and numerical answers, which might be used in open-world problems.
We argue that using LLMs to improve themselves is limited; interactions between LLMs and environment are essential incremental  learning. 

There are two typical ways for incremental  learning:   \textit{offline incremental training} and \textit{online in-context learning}.\textbf{ Offline incremental training} in language models involves  \replace{fine-tuning}{training} the system with new sets of annotated data, allowing for some level of adaptability\add{, such as supervised fine-tuning~\cite{ouyang2022training} and reinforcement learning with human feedback~\cite{christiano2017deep, ziegler2019fine}}. However, this approach is fraught with several limitations. First, it suffers from a lagging nature, meaning there is a delay between the emergence of new information and the model's update to include it, rendering the model less reliable for immediate or evolving scenarios. Second, the method is generally non-personalized; it updates the model based on broad data sets rather than tailoring to individual user preferences or specific contextual needs. This method's high computational cost and static nature post-update make it inflexible for adapting to real-time changes or diverse user requirements. 

Another approach in the realm of incremental learning is known as '\textbf{online in-context learning}'. In this setup, a LLM typically interacts with an AI agent or connects to either offline or online knowledge sources. For instance, the model might retrieve information from a given knowledge base or from web content, a.k.a, Retrieval-Augmented Generation~(RAG)~\cite{izacard2022few, wang2023shall, lewis2020retrieval}. Such methods often occur within the paradigm of in-context learning~(ICL)~\cite{brown2020language,liu2021makes}. However, a significant limitation of this approach is its lack of knowledge persistency. When the session changes, the learned information is not retained, leading to a loss of any updates or learning that took place.

To address the shortcomings of both 'Offline Incremental Training' and 'Online In-Context Learning,' we introduce a cutting-edge paradigm, 'Online Training using External Interactions.' This new approach offers the benefits of persistent model updates and real-time learning. Unlike traditional methods, it necessitates external interactions during the learning process, such as interfacing with AI agents or connecting to offline or online knowledge sources. These capabilities allow the model to continually adapt and stay updated, effectively addressing the limitations seen in previous frameworks. \add{Furthermore, this innovative human-computer interaction paradigm presents a unique opportunity for ordinary users to modify LLMs themselves, thereby shifting from a developer-owned model to a more user-centric approach. This paradigm could mark the beginning of a broader practical application of LLMs in the everyday lives of an increasing number of people.}


The key contributions of this work are as follows:  

\begin{itemize}
    \item  \add{ We propose a novel paradigm for human interaction with language models, shifting from static to adaptable models. Instead of humans adapting to fixed model parameters, we introduce models that evolve in response to human input, fostering a dynamic and sometimes reciprocal relationship. }
    \item  The classification of interaction paradigm and proposal of a novel  paradigm: Online Training using External Interactions, which is a user-friendly incremental learning methodology.  
\end{itemize}

\begin{table}[tbp]
\centering
\caption{Comparison of interaction paradigms. Online and offline refer to whether the model is serving, while training and parameter-invariant and parameter-variant indicate whether the parameter of the model changed. To assess and compare various interaction approaches, we focus on five key attributes: 1) Knowledge Persistency, which indicates if updated information remains accessible across different sessions; 2) Flexibility, evaluating if \replace{Large Language Models (LLMs)}{LLMs} become static post-training; 3) Efficient Update, gauging the time and computational costs involved in model updates; 4) Knowledge Timeliness, assessing if the model’s information is current; and 5) Knowledge Quality, which verifies the accuracy and reliability of the model's information. In the context of traditional training, LLMs are fine-tuned using a specific set of annotated data.}
\label{tab: Interaction Paradigm Comparison}
\scalebox{0.8}{
\begin{tabular}{c|c|ccccc}
\hline
Paradigm & Methodology & \textbf{\thead{Knowledge \\ Persistency}} & \textbf{\thead{Flexibility}} & \textbf{\thead{Efficient \\ Update}} & \textbf{\thead{Knowledge \\ Timeliness}} & \textbf{\thead{Knowledge \\ Quality}} \\
\hline
\thead{Offline \\Parameter-Variant} & \thead{Traditional Training~\cite{brown2020language, schick2022peer, malmi2022text}} & \checkmark & & & & \checkmark \\ 
\hline
\multirow{3}*{\thead{Online \\ Parameter-Invariant}} & \thead{\replace{Retrieval-Augmented Generation\\~(RAG)}{Retrieval-based methods}~\cite{ izacard2022few, wang2023shall, lewis2020retrieval}} & & \checkmark & \checkmark & & \\
~ & \thead{\replace{In-context Learning}{Prompt-based methods}~\cite{brown2020language,liu2021makes}} & & \checkmark & \checkmark & & \\
~ & \thead{\replace{Tool-Enhanced}{Tool-based methods}~\cite{langchain2023, haystack2023,hfTextGenerationInference2023, vLLM2023, yang2023skypilot, FastChat2023,qin2023toolllm}} & & & & \checkmark & \\
\hline
\textbf{\thead{Online \\ Parameter-Variant}} & \textbf{\thead{Online Training \\ using External Interactions}} & \checkmark & \checkmark & \checkmark & \checkmark & \checkmark \\
\hline
\end{tabular}}
\end{table}

\section{Related Work and Motivation}

\add{From the users' perspective, there are currently two well-known interaction paradigms with LLMs: offline parameter-variant and online parameter-invariant. In this section, we first introduce these two prevalent user interaction paradigms and their applications, along with their applications, and detail their respective advantages and disadvantages in Sec.~\ref{sec:related work}. Following this, in comparison to existing works, we outline the research objectives and motivation behind the proposed novel interaction paradigm in Sec.~\ref{sec:motivation}}.

%

\subsection{Related Work}
\label{sec:related work}

\paragraph{\add{{\textbf{Offline parameter-variant paradigm}}}} The offline parameter-variant paradigm is the most commonly used interaction paradigm, wherein models are updated during periods of non-service. This paradigm comprises methods that are trained on a given labeled dataset. The dataset can be compiled solely by humans, as suggested by \add{Ouyang et al. (2022)~\cite{ouyang2022training}, which represents a conventional method of model training, specifically referred to as supervised fine-tuning~(SFT)}. Alternatively, the process can be assisted by a retriever, as explored by ~\cite{hu2023survey,liu2022relational,lu2021kelm}. Interaction with external knowledge during training can enhance the model's representation by integrating a larger volume of factual knowledge. \add{For developers, offline parameter-variant methods are the most reliable and effective options for training a language model from scratch or for model updates due to their 'once and for all' characteristic, leading to high knowledge persistency and quality. However, from the user's perspective, the entire model training process can be complex, inflexible and time-consuming, ranging from data collection to computing resource configuration and model training. }

\paragraph{\add{\textbf{Online parameter-invariant paradigm}}} For online parameter-invariant paradigm, there are three kinds of techniques, \textit{retrieval-based methods}~\cite{izacard2022few, wang2023shall, lewis2020retrieval}, \textit{prompt-based methods}~\cite{brown2020language,liu2021makes}, and \textit{tool-based methods}~\cite{langchain2023, haystack2023,hfTextGenerationInference2023, vLLM2023, yang2023skypilot, FastChat2023,qin2023toolllm}. RAG~\cite{lewis2020retrieval} is a representation of \textit{retrieval-based methods}, which emphasizes the use of external knowledge sources to augment language models during inference time. Interaction with the knowledge base during inference can aid the language model in generating more precise, contextually relevant, and informed responses by dynamically leveraging external knowledge sources based on the specific input or query at hand. RAG amalgamates pre-trained parametric and non-parametric memory for language generation. Designed to enhance the factual accuracy of dialogue agents, it aims to mitigate the issue of knowledge hallucination. Whereas, its performance heavily relies not only on the quality of the knowledge but also on the effectiveness of the retrieval method. 

The primary objective of the \textit{prompt-based method} is to sustain real-time, continuous engagement, making it ideal for application scenarios such as dialogue systems, real-time translations, and multi-round question answering. This iterative interaction process allows the model's output to incrementally adjust to meet user demands. Typically, this interaction paradigm does not modify the model's parameters during the interaction, instead necessitating users to incessantly input or revise prompts to draw more meaningful responses from the language model. In-context learning ~\cite{brown2020language} is a method of prompt-based, which operates without access to any external memory or knowledge beyond its pre-training phase. Consequently, while generating responses, it primarily relies on the immediate context of the conversation or task for information. Consequently, conversations can become rigid and labor-intensive due to the necessity for prompt engineering or dialogue engineering. The drawbacks of this interaction pattern are the inefficiency of proper prompts construction and failures in non-textual tasks.  

\add{By fragmenting a downstream task into multiple steps, \textit{tool-based methods} can assign specific stages to  external tools or APIs, such as those specializing in mathematical computations, web searches, image generation, and etc.} Typically, tasks that emphasize fidelity and accuracy, such as real facts, complex mathematical operations, and tasks that transcend the LLM training corpus including up-to-date knowledge, low-resource languages, and image generation, are more effectively resolved using external tools than LLMs. ToolLLM ~\cite{qin2023toolllm} is a well-known tool-enhanced method. It constitutes a comprehensive framework for tool-use, offering tangible tools and components for LLMs. It is specifically engineered to empower LLMs to execute higher-level tasks, such as adhering to human instructions for utilizing external tools (APIs). However, it suffers from challenges related to invoking tools at the appropriate time and determining the most suitable tool to utilize.

\paragraph{\add{\textbf{Pros and Cons of Existing Paradigms}}} \add{The two existing User-LLM interaction paradigms are extensively utilized in both research and practical applications. Offline parameter-variant approaches excel in knowledge persistency and quality, a result of the substantial engineering effort required before model training. This includes thorough data collection and meticulous training configuration. However, such a workload leads to inflexibility and high costs in model updates. Moreover, these methods inherently lack timeliness in knowledge updates due to their demanding workload. On the other hand, online parameter-invariant methods enable the enhancement of trained LLMs without extra training costs through prompting. However, as observed in previous research~\cite{liu2023lost}, the efficacy of prompting is not consistently positive. It may significantly degrade, especially if the relevant information's position varies in long context, even in models designed for long-context scenarios. Additionally, online parameter-invariant methods, like retrieval-based and tool-based approaches, often require extensive systems support, such as external databases, posing a burden for system development and sharing. Furthermore, the integrated knowledge in these methods has a short effective lifespan, expiring at the session's end or upon context removal. Overall, the pros and cons of existing paradigms are listed in Table~\ref{tab: Interaction Paradigm Comparison}}.

\subsection{Motivation}
\label{sec:motivation}

\paragraph{\add{\textbf{Motivation}}} \add{In the context of knowledge persistence, the offline parameter-variant paradigm can be likened to the human brain's long-term memory, which necessitates extensive training and preparation for embedding specific knowledge. Conversely, the online parameter-invariant paradigm resembles short-term memory, where knowledge or skills can be rapidly acquired through quick learning but are not retained in the model for an extended period. This analogy highlights the strengths and weaknesses of these existing interaction paradigms. Motivated by this, we propose an intermediate interaction paradigm that amalgamates the benefits of both: the 'Online Parameter-Variant' method. This novel approach aims to reduce training costs compared to offline parameter-variant methods while offering more robust improvements than the online parameter-invariant paradigm. The 'Online Parameter-Variant' method, which is grounded in model training, focuses on several key metrics: knowledge persistency, flexibility, efficient updating, knowledge timeliness, and superior knowledge quality relative to the previous paradigms.}

\section{User Interface\add{: online training using external interactions}}

\add{We will first introduce the overall design in Sec. 3.1 which consists of three interactions. These three interactions are detailed in Sec. 3.2, Sec. 3.3 and Sec. 3.4.}

\subsection{\add{Overall design}}

\subsubsection{\add{Philosophy}}
To address these challenges, we introduce \replace{an innovative}{a new} interactive interface that facilitates user engagement with LLMs through conversational interactions while concurrently enabling fine-tuning through natural language instructions. The proposed system allows users to engage in conversations with a LLMs while providing specific instructions to trigger immediate fine-tuning. Users can seamlessly trigger the training process by employing natural language prompts preceded by "\textbf{\textit{[Learn]}}," like \textbf{\textit{"[Learn] I wish you could fetch more news on environmental pollution,"}} within an interface resembling a chat, thereby commencing training grounded on network-sourced information.
\delete{We present this approach as online learning with interaction.}

Upon receiving the triggering signal, our system will comprehend the user's intended meaning and initiate distinct learning processes accordingly. After the training is completed, the newly enhanced model, enriched with incremental knowledge, will immediately replace the preceding model and seamlessly resume the ongoing conversation with users. 

\delete{Online learning with interaction includes three distinctive learning functionalities:}

\subsubsection{\add{The three interactions}}
\add{Online Training using External Interactions introduces three unique learning functionalities that form a comprehensive and versatile toolkit for interactive model training:}
\textbf{Instruction-Guided Learning},\textbf{ Document-Driven Learning}, and \textbf{Web Search-Enabled Learning}. 
\delete{Together, these functionalities constitute a multifaceted and comprehensive toolkit, empowering users in the interactive refinement and adaptation of language models. By employing \textit{self-instruction}, \textit{instruction backtranslation} and \textit{online search augmentation} for data generation, our functionalities enable users to actively shape and customize their models based on specific instructions, documents, and real-time web data, fostering adaptability and personalization in various application domains. Combining dynamic chat-based interactions with real-time training capabilities, our framework fosters a user-centric and continuous improvement process. }

\add{\textbf{Instruction-Guided Learning} serves as a soft knowledge source, leveraging conversational interfaces like \textit{ChatGPT}~\footnote{https://chat.openai.com} to facilitate human-like, adaptive responses. This functionality is particularly powerful for nuanced or subjective queries where human-like interpretation and flexibility are required.

On the other hand, \textbf{Document-Driven Learning} and \textbf{Web Search-Enabled Learning} act as hard knowledge sources. \textbf{Document-Driven Learning} relies on offline sources, allowing for quality-controlled, curated information to be used in model training. This is particularly advantageous for tasks that require authoritative or highly reliable information. \textbf{Web Search-Enabled Learning} utilizes online data, offering the advantage of real-time information retrieval. While this allows the model to stay current, it can sometimes introduce bias or less reliable data into the training set.}

\add{By using \textit{self-instruct~\cite{wang2022self}}, \textit{instruction backtranslation~\cite{li2023self}}, and \textit{online search augmentation}, these functionalities allow for a high degree of customizability and adaptability. They empower users to shape their models according to specific instructions, documents, or real-time web data, thus bridging the gap between static, pre-trained models and dynamic, personalized user needs.
Together, these functionalities make our framework not only versatile but also user-centric, enabling continuous improvement and adaptability across various application domains. This effectively highlights the complementary nature of the three functionalities in offering different sources and reliability of knowledge, serving to create a balanced, comprehensive approach for online learning with interaction.}

\subsubsection{\add{Content Moderation Control}}
\add{Content moderation is crucial for maintaining the integrity of LLMs. To ensure effective moderation and reduce the risk of generating biased, toxic, or unethical content, the proposed interface utilizes two primary strategies: Prevention and Feedback. First, we employ an external interface specifically designed to monitor and address content moderation issues during training. Second, we integrate a user feedback mechanism, enabling users to contribute to the moderation process through their interactions and observations. For the prevention aspect, all data used for model updates will undergo rigorous scrutiny, filtering out any inappropriate content to ensure that sources, whether local documents or Internet-based, are suitable for LLM moderation. As for the feedback component, a 'feedback' button is available to users, allowing them to report biases or any unsatisfactory elements of the response. When used, the model initiates an updating process, making corrections through a pre-defined mechanism. Interestingly, this feedback mechanism can be viewed as a form of online version reinforcement learning with human feedback (online RLHF)~\cite{ouyang2022training}, aimed at aligning LLMs with actual user values and accentuating the personalized aspect of the proposed method.}

\subsubsection{\add{Benefits and potential.}}

\paragraph{\add{\textbf{Lifelong Learning}}}\delete{~\cite{Biesialska_2020}}
The concept of lifelong learning\add{~\cite{Biesialska_2020}} refers to the ability of a system, in this case, a language model, to continuously acquire and integrate new knowledge and skills throughout its existence. Unlike traditional training methods that rely on static datasets, our approach leverages continuous user interactions to adapt and evolve the model over time. This enables lifelong learning, where the model can continuously improve and stay relevant to the user's changing needs and interests.
\paragraph{\textbf{Personalization~\cite{chen2023large}}}
Through our interactive training mode, users have the power to customize the language model to their liking. This customization extends beyond simple prompt-based instructions and allows users to fine-tune the model's behavior to suit their unique requirements. This level of personalization results in models that are highly specialized and context-aware, enhancing their utility for specific tasks and domains.
\paragraph{\textbf{Accessibility}}
The heart of our approach lies in allowing users to engage in natural language conversations with the language model. This familiarity with conversational interactions makes it accessible to a wide range of users, regardless of their technical background. Users don't need to learn complex programming or command syntax; they simply converse as they would with another person.
\paragraph{\textbf{User Empowerment}}
Users are in control of the training process. They can decide when and how to fine-tune the model based on their needs. This sense of empowerment fosters a feeling of ownership over the model, enhancing user engagement and motivation to participate in the training process. Additionally users can track the progress of their customized model over time. They can see how their interactions and commands have shaped the model's behavior, providing a sense of achievement and transparency in the training process.

\subsection{Instruction-Guided Learning}
Instruction-Guided Learning constitutes a fundamental component of our interactive language model fine-tuning system, enabling users to impart specific directives to the model regarding information retention and contextual understanding during ongoing conversational interactions. This method is particularly instrumental in customizing the model's responses to align with the user's unique requirements and preferences.

\begin{figure}[htb]
    \centering
      \includegraphics[width=1.0\textwidth]{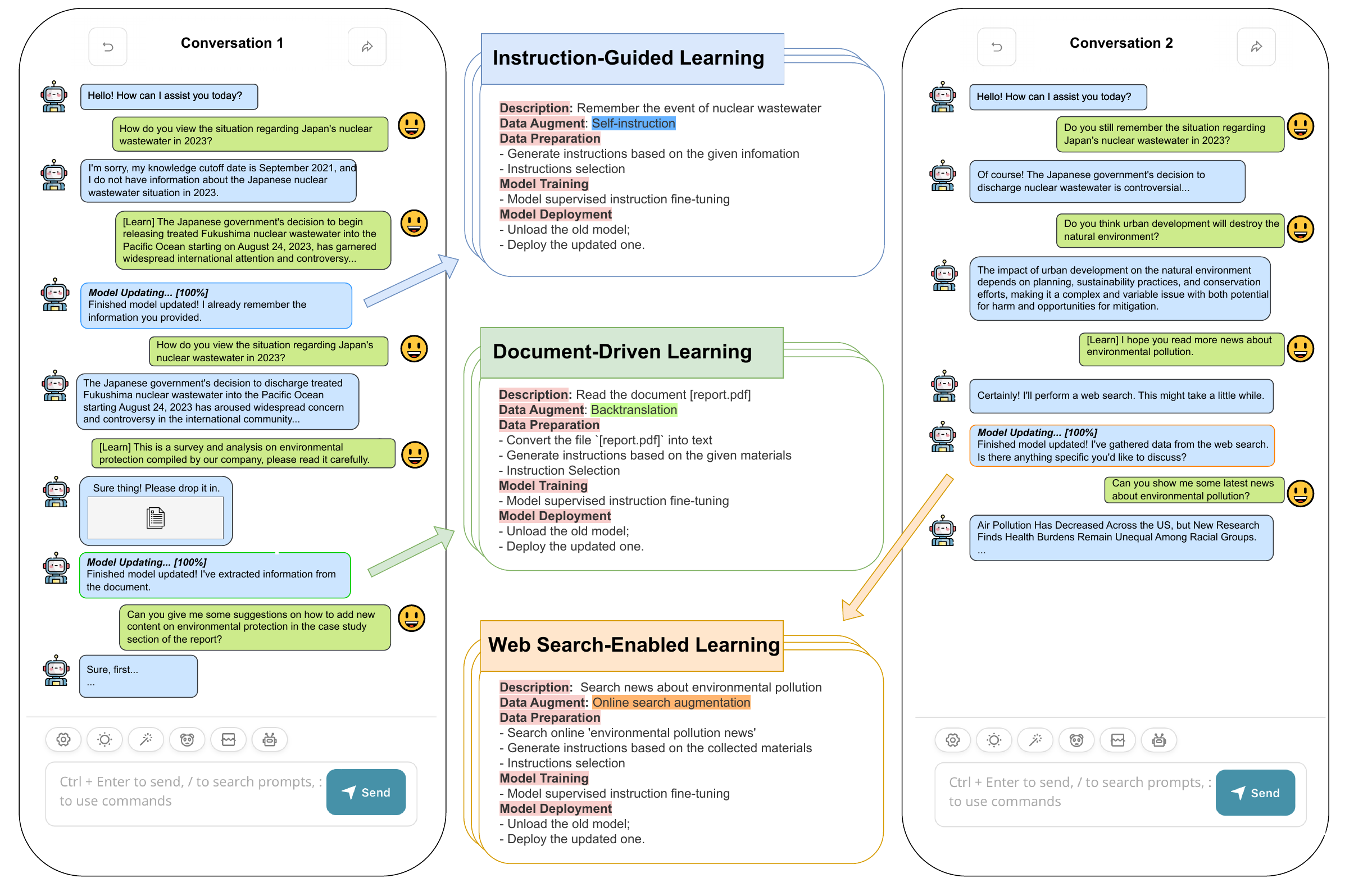}
    \caption{\textbf{The figure depicts the manner in which dialogues are conducted between LLM and user within our interactive mode. Notably, users issue distinct directives, each leading to the trigger of three distinct training processes. Furthermore, the figure underscores the model's ability to retain knowledge acquired during prior conversational session, even when transitioning across different conversation sessions.}}
    \label{fig:llmfactory1}
\end{figure}

Within the Instruction-Guided Learning method, users are granted the capability to issue explicit instructions to the language model. These instructions may pertain to what facts or details the model should remember or any specific information that should be considered during the discourse. Users can convey their instructions in a natural language format, making it an intuitive and user-friendly process. These directives will be transmuted into trainable data via the \replace{self-instruction approach\cite{wang2022self}}{self-instruct approach~\cite{wang2022self}}, after which we will proceed to iteratively enhance our model using the data thus generated.

After receiving user instructions, the model promptly incorporates the provided information into its understanding of the ongoing conversation. This entails the identification and retention of salient details and context specified by the user. The model's responses are then guided by the personalized context, resulting in responses that align closely with the user's directives.

As the conversation unfolds, the model continuously adapts its responses based on the instructions and context provided by the user. This iterative adaptation process allows the model to tailor its responses, ensuring that it adheres to the user's preferences and maintains a coherent and contextually relevant dialogue. Consequently, the user experiences a personalized and highly responsive conversational interface.

Instruction-Guided Learning offers users a powerful means to personalize the language model's behavior and responses in a conversational setting. By issuing explicit directives, users can shape the model's understanding and context, thereby tailoring its responses to their unique requirements. This method enhances the utility of language models across various applications, including personal virtual assistants, domain-specific chatbots, and tailored information retrieval systems, making them versatile tools for a diverse range of user needs.

\begin{figure}[htb]
    \centering
      \includegraphics[width=0.9\textwidth]{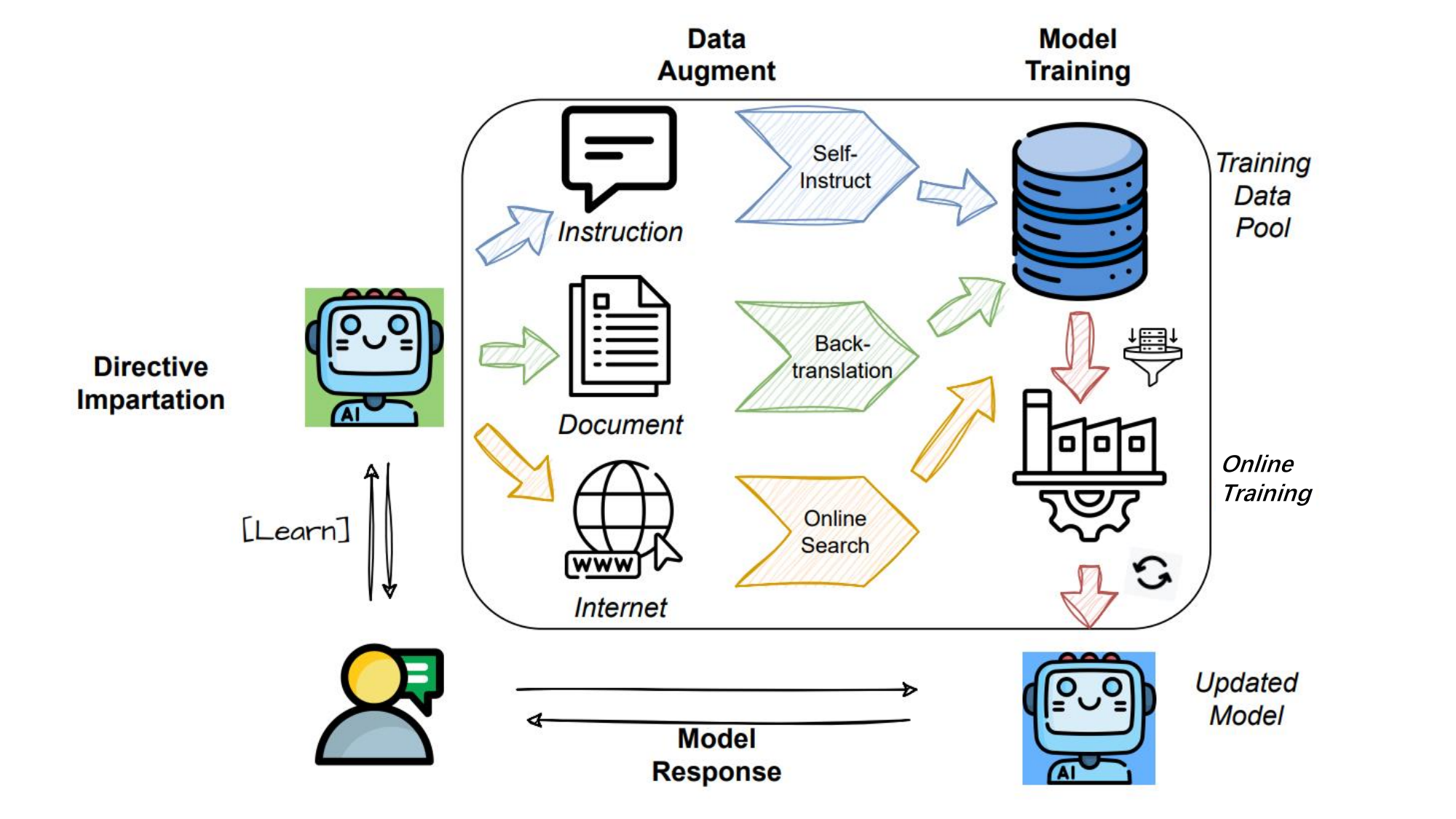}
    \caption{\textbf{This figure delineates our comprehensive workflow of chat-based online training. During the interaction between the user and the model, the user issues learning instructions to trigger the learning process. Three different learning methods correspond to three data augmentation techniques with the generated data as input to train new model. Then new model replace the old one seamlessly, allowing the user to continue the conversation.}}
    \label{fig:llmfactory2}
\end{figure}

\subsection{Document-Driven Learning}

Document-Driven Learning constitutes a pivotal facet of our interactive language model fine-tuning system, offering users the capability to enrich the model's knowledge base with structured and specialized information. \delete{\cite{guu2020retrieval}}This mode is particularly suited for users who seek to imbue the model with domain-specific expertise or train it on authoritative documents, academic texts, or specialized knowledge sources.

Users initiate the Document-Driven Learning method by selecting and uploading documents relevant to their specific area of interest or domain. These documents may encompass scholarly articles, technical manuals, legal documents, or any textual resources germane to the subject matter. The system accepts a variety of file formats, including PDFs, text documents, and web links.

Upon document submission, the system undertakes a comprehensive preprocessing and transformation procedure.\delete{\cite{li2023self}} Document-derived data produced through the utilization of Instruction Backtranslation\add{~\cite{li2023self}} will be meticulously curated for high-quality training purposes. Through iterative fine-tuning, the model adapts to the new information derived from the uploaded documents. It learns to contextualize the content, recognize domain-specific terminology, and develop a deeper understanding of the subject matter. Consequently, the model's responses become more nuanced and contextually relevant when engaging in discussions related to the uploaded documents or the associated domain.

Document-Driven Learning represents a potent mechanism for users to imbue language models with domain-specific knowledge and expertise. By leveraging structured textual resources, users can enhance the model's contextual understanding and its capacity to provide informed responses within specialized domains. This method extends the utility of language models across a wide array of professional and academic applications, enabling them to serve as versatile and knowledgeable conversational partners.
\subsection{Web Search-Enabled Learning}
The integration of Web Search-Enabled Learning constitutes the third facet within the framework of our interactive language model fine-tuning system, affording users the capability to harness the vast knowledge repository of the internet to augment the model's understanding and responsiveness. This method is particularly valuable for users seeking real-time information, staying updated on current events, or training the model on a dynamic and ever-evolving knowledge landscape.

Upon receiving user search instructions, the system promptly conducts web searches using well-established search engines and APIs. The retrieved web content, which may include news articles, blog posts, research papers, and other relevant sources, is then subjected to information extraction and summarization processes to distill the key insights and facts. The extracted information from web searches serves as a valuable source of training data for the model~\cite{wang2022self}. During the subsequent fine-tuning phase, the model is exposed to the insights obtained from web searches.

An inherent advantage of Web Search-Enabled Learning is the model's adaptability to real-time information. As the web content evolves, the model continuously adapts to the dynamic knowledge landscape, ensuring that its responses remain up-to-date and accurate in the context of the ongoing conversation. This real-time adaptation is particularly advantageous for users seeking the latest information and insights.

The knowledge derived from web searches becomes an integral part of the model's memory, enriching its understanding of contemporary topics and factual information. This knowledge integration ensures that the model remains a reliable source of current events, trending topics, and dynamic knowledge domains over time.

Web Search-Enabled Learning empowers users to leverage the extensive resources of the internet to enhance the language model's knowledge and responsiveness. By instructing the system to retrieve real-time information, users ensure that the model remains current and up-to-date, making it a valuable resource for information retrieval, news updates, and dynamic knowledge domains. This method extends the utility of language models to domains requiring real-time knowledge integration and adaptation, making them versatile tools for a wide array of applications, including news summarization, trend analysis, and current event discussion.

\add{
\section{Application: a Case Study on Tool Learning}
This section is dedicated to evaluating the effectiveness and efficiency of the proposed novel interaction paradigm, termed \textit{Online Training using External Interactions}, abbreviated as \textit{\textbf{Online Training (OT)}} in this section.

\subsection{Problem setting}

In this task, we assume that the user's objective is to train a LLM to effectively utilize external tools~\cite{sun2023moss, qin2023toolllm}. To achieve this goal, we adopt the tool invocation data format outlined in Sun et al. (2023)~\cite{sun2023moss}, as demonstrated in Appendix~\ref{appendix: experiment details}. We assess the model's accuracy in invoking the correct plugin and its corresponding inputs when presented with multiple APIs for various questions.

As illustrated in Figure~\ref{fig:taskA workflow}, we employ two baseline methods: the prompt-based method (abbreviated as \textit{Prompt}) and the full-parameter training method (abbreviated as \textit{Full-SFT}). In the \textit{Prompt} method, we use the base model (Llama2-7b-chat~\cite{touvron2023llama}) to generate answers with few-shot prompts~\cite{wang2020generalizing} listed in the context. Conversely, the \textit{Full-SFT} method involves leveraging external annotated training data to train the model, subsequently using the same few-shot prompts as in the \textit{Prompt} method for the test set.
And, the proposed approach, online-training~(OT), involves generating corresponding training data based on user training instructions, then filtering out low-quality, including toxic or biased, data to train the original model. Subsequently, a prompt-based approach is employed to generate answers for the test set. We select three tools from previous research~\cite{qin2023tool}, randomly choosing 300 data points for the test set and an additional 6k data points for the training dataset for the Full-SFT method

\begin{figure}[htb]
\centering
\begin{minipage}[t]{0.48\textwidth}
\centering
\includegraphics[width=\textwidth]{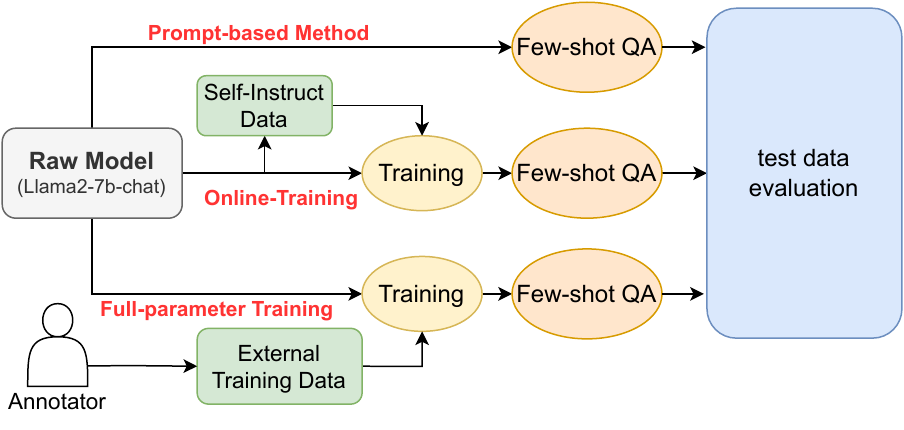}
\caption{Overview of the experimental design}
\label{fig:taskA workflow}
\end{minipage}
\begin{minipage}[t]{0.48\textwidth}
\includegraphics[width=0.9\textwidth]{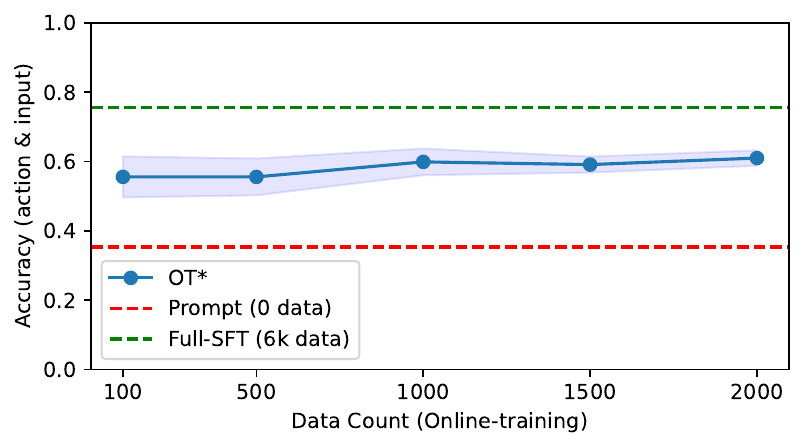}
\caption{The results of the experiment, where the symbol \textbf{*} refers to the average of three experiments with random seed.}
\label{fig:taskA result}
\centering
\end{minipage}
\end{figure}

\subsection{Result Analysis}

The result are shown in Figure~\ref{fig:taskA result}, where the x-coordinate indicates the data utilized by the \textit{online-training} methods, and the y-coordinate shows the accuracy of each method using external tools. Accuracy is measured by counting instances of correctness in both the action (correct tool selection) and input (accurate tool parameters generation). The \textit{Full-SFT} method utilizes 6,000 labeled in-domain data points for model training. In comparison, the performance of the OT method is evaluated using between 100 and 2,000 model-generated data points for training. It is evident that using a single round of model generation~\footnote{Approximately 100 valid data points can be obtained from a single GPT-4 API call.} can achieve almost double the improvement over the vanilla model in the tool learning task, increasing from nearly 30\% to 50\%. However, the performance of online-training trained on 100 data points compared to 2000 data points appears similar. This is attributed to data distribution misalignment, which is still considered acceptable.

\begin{table}[htb]
\centering
\begin{tabular}{lccc}
\toprule
Method & Prompt & OT (0.1k) & SFT (6k) \\
\midrule
Accuracy & 0.35 & 0.56 & 0.76 \\
Train (time) 
& / & $2 mins$ & $40 mins$ \\
Inference (time)
& $2 mins$ & $2 mins$ & $2 mins$ \\
\bottomrule
\end{tabular}
\caption{Analysis of experiment duration: the time expended during the training process and the inference time for the test set are detailed.}
\label{tab: overall result}
\end{table}

Table~\ref{tab: overall result} presents the analysis of experiment duration, detailing the time expenditures for both the training and inference processes. It becomes evident that the online-training method effectively amalgamates the benefits of both online parameter-invariant and offline parameter-variant methods, as demonstrated by its reduced training cost and heightened effectiveness on the test set. 

\section{Discussion}

This section, drawing upon the experiments and related works, delves into some concerned issues and potential challenges associated with the proposed method. Also, future possibilities in the development of User-LLM interaction paradigms are introduced.

\subsection{In-context Learning or Fine-Tuning?} 
There is always a question regarding language model downstream adaptation: should we opt for in-context learning or model fine-tuning for the continuous learning of trained LLMs? Both approaches have garnered considerable interest from researchers and users alike. As previously discussed, each paradigm has its strengths and weaknesses, depending on the scenario, and they are, in fact, not mutually exclusive. Moreover, increasing research~\cite{ovadia2023finetuning, dai2023gpt} is focusing on the relationship between ICL and Fine-Tuning. Dai et al. (2023)~\cite{dai2023gpt} found that ICL behaves similarly to explicit fine-tuning at the prediction level, representation level, and attention behavior level. In light of these findings, we propose a novel interaction paradigm that bridges the gap between these two existing approaches. This method involves injecting knowledge directly into the parameters, rather than solely in the context, thereby enhancing its persistency and robustness.

\textbf{Scalability}: One of the prominent advantages of the proposed method over ICL is its superior transferability and compositionality. For example, one could  prepare specific training data for  each learning job and  then later decide which training data to be combined for final application. Such a combination could be done without extra inference cost as the increase of learning jobs does not affect the inference cost.  Also thanks to the efficient compositionality, our approach could have a better capacity to deal with a larger-scale training, which benefits the scalability.

\textbf{Inference Efficiency}
Our method stands out for its superior inference efficiency when compared to Retrieve and Generate (RAG) or In-Context Learning (ICL) strategies. Both RAG and ICL often result in significantly longer input prompts, which in turn leads to an increase in computational cost - a cost that grows quadratically with the length of the input. Although the training cost of our approach might be higher than that of RAG or ICL, it's important to note that this is a one-time expense related to the training phase, and remains constant regardless of the number of requests. In contrast, the cost of inference increases linearly with the number of requests, making our method more efficient in the long run. Moreover, unlike approaches that require large-scale databases or additional plugins, our method incorporates knowledge directly into the model through online training, thereby eliminating the need for external dependencies. This not only simplifies the process, but also enhances deployment readiness and operation efficiency. 

\subsection{Challenges}
Despite the potential of the online training method, it faces several challenges:

\begin{itemize}
    \item \textbf{Knowledge Injection and Overfitting:} Ovadia et al. (2023)~\cite{ovadia2023finetuning} noted that LLMs often struggle to assimilate new factual information through fine-tuning. A key challenge is effectively injecting necessary knowledge into LLMs within a user-acceptable timeframe to enhance user experience. Rather than relying on a high number of training epochs, which may cause models to overfit by repeatedly training on the same data, our approach increases data diversity. This aligns with user requirements and ensures model generalizability and knowledge acquisition.
    
    \item \textbf{Knowledge Persistency:} Maintaining the knowledge persistency in LLMs is a crucial aspect of our proposed system. Unlike ICL-type knowledge persistency, which may involve storing information on disk, our parameter-variant methods embed knowledge directly into the LLMs' parameters. This approach ensures long-lasting knowledge retention, akin to pre-trained knowledge.
    
    \item \textbf{Concurrency in LLMs Deployment:} The online-training interaction paradigm we propose has similar deployment costs to conventional methods. It can be trained into a specific set of parameters, such as using LoRA, which offers the flexibility to load or unload specific training for users. This ensures privacy and scalability without significantly increasing deployment demands.
\end{itemize}

}

\section{Conclusion}
\add{In this paper, we introduce a novel interaction paradigm, online parameter-variant, and a new method, online learning using external interactions. This approach focuses on explicit model fine-tuning and instant responses to natural language instructions via a user-friendly interface. As for the future direction, we aim to break down more restrictions on users’ utilization of LLMs and bring more engaging and beneficial model human-computer interactions to users, leveraging the 'Online Training using External Interactions' paradigm.}


\section*{Acknowledgement}
This work is supported by the Shenzhen Science and Technology Program (JCYJ20220818103001002), Shenzhen Doctoral Startup Funding (RCBS20221008093330065), and Tianyuan Fund for Mathematics of National Natural Science Foundation of China (NSFC) (12326608).

\bibliographystyle{unsrt}
\bibliography{base}

\appendix

\add{
\section{Experiment Details}
\label{appendix: experiment details}

For all experiments conducted in this study, we utilized four A100 GPUs, each equipped with 80GB of memory. The learning rate for SFT was set to 2e-06, while for online training, it was established at 2e-5. Following the guidelines in Tirumala et al. (2022)~\cite{tirumala2022memorization}, we set the training batch epoch at 10 for OT and at 2 for SFT.

Additionally, the data format for tool invocation is exemplified as follows:
}

\vbox{
\begin{lstlisting}[language=CIL]
Human: Can you provide a weather forecast for Rio de Janeiro, Brazil for the upcoming weekend?
GPT: Thought: I need to use the forecast_weather API to get the weather forecast for Rio de Janeiro, Brazil for the upcoming weekend.
    Action: weather.forecast_weather
    Action Input: {"location": "Rio de Janeiro, Brazil", "days": 2}
\end{lstlisting}
}

\end{document}